\documentclass{article}

\usepackage{spconf,amsmath,graphicx}
\usepackage{multirow}
\usepackage{indentfirst}
\usepackage{enumitem}
\usepackage[numbers,sort&compress]{natbib}
\usepackage[hidelinks]{hyperref}

\def \f1 {$f_1$-score }

\title{Transfer Learning in ECG Diagnosis: Is It Effective?}
\name{Cuong V. Nguyen$^1$ and  Cuong D.Do$^{1,2}
$\sthanks{Corresponding author.
Source code is available at \texttt{\href{https://github.com/cuongvng/transfer-learning-ecg-diagnosis}{github.com/cuongvng/transfer-learning-ecg-diagnosis}}.}
} 

\address{
$^1$ College of Engineering and Computer Science, VinUniversity, Hanoi, Vietnam \\
$^2$ VinUni-Illinois Smart Health Center, VinUniversity, Hanoi, Vietnam \\
\texttt{\{cuong.nv,\ cuong.dd\}@vinuni.edu.vn}
}

\begin{document}

%
\pagestyle{plain}
\pagenumbering{arabic}

\maketitle
\begin{abstract}

The adoption of deep learning in ECG diagnosis is often hindered by the scarcity of large, well-labeled datasets in real-world scenarios, leading to the use of transfer learning to leverage features learned from larger datasets.
Yet the prevailing assumption that transfer learning consistently outperforms training from scratch has never been systematically validated.
In this study, we conduct the first extensive empirical study on the effectiveness of transfer learning in multi-label ECG classification, by investigating comparing the fine-tuning performance with that of training from scratch, covering a variety of ECG datasets and deep neural networks.
We confirm that fine-tuning is the preferable choice for small downstream datasets; however, when the dataset is sufficiently large, training from scratch can achieve comparable performance, albeit requiring a longer training time to catch up.
Furthermore, we find that transfer learning exhibits better compatibility with convolutional neural networks than with recurrent neural networks, which are the two most prevalent architectures for time-series ECG applications.
Our results underscore the importance of transfer learning in ECG diagnosis, yet depending on the amount of available data, researchers may opt not to use it, considering the non-negligible cost associated with pre-training.

\end{abstract}


\section{Introduction}

Electrocardiogram (ECG) signals play a critical role in the early detection and diagnosis of cardiovascular diseases. 
The integration of automatic ECG interpretation, fueled by digitization and deep learning, has demonstrated performance on par with cardiologists \cite{hannun2019cardiologist, ribeiro2020automatic}.
A major challenge to wide-scale adaptation of deep learning to ECG diagnosis is the lack of large-scale, high-quality labeled datasets in most real-world scenarios, due to prohibitive collection and annotation costs.
To overcome this challenge, transfer learning is commonly employed, where features and parameters learned from a large dataset are reused and fine-tuned on a typically smaller, new dataset.
This technique has been adapted from computer vision \cite{finetuning1, kornblith2019better, finetuning2, headretraining1, headretraining2} to the ECG domain.
Some studies have applied transfer learning to classify ECG arrhythmia by borrowing pre-trained weights on 2-D ImageNet \cite{imagenet}, after transforming 1-D ECG signals to 2-D representations.
For example, Salem et al. \cite{salem2018ecg} generated 2-D spectrograms from ECG using Fourier Transform and applied pre-trained weights of DenseNet \cite{densenet} on ImageNet \cite{imagenet} to detect ventricular fibrillation, atrial fibrillation, and ST-changes.
Tadesse et al. \cite{tadesse2019cardiovascular} also employed spectrograms and leveraged pre-trained inception-v3 GoogLeNet \cite{szegedy2014going} to diagnose cardiovascular diseases.
Gajendran et al. \cite{gajendran2021ecg} and Venton et al. \cite{venton2020signal} leveraged scalogram for 2-D conversion and reusing convolutional neural networks (CNNs) \cite{densenet,szegedy2014going,vgg,resnet,darknet13,tan2019efficientnet} pre-trained on ImageNet to classify ECG records.
Zhang et al. \cite{zhang2021heartbeats} applied Hilbert Transform and Wigner-Ville distribution \cite{sultan2017ecg, dhok2020automated} to convert signals to 2-D then using pre-trained ResNet101 \cite{resnet} to build their classifiers.

\begin{table*}[ht]
  \centering
  \caption{Datasets used in this work.} \vspace{2mm}
  \label{table:dataset_summary}
  \begin{tabular}{lcccc}
      \hline
    Dataset & Labels used & Samples & Training samples & Testing samples \\
    \hline
    PTB-XL \cite{ptbxl_dataset} & 5 & 21,837 & 17,441 & 2,203 \\
    CPSC2018 \cite{cpsc2018} & 9 & 6,877 & 4,603 & 2,268 \\
    Georgia \cite{georgiadataset} & 10 & 10,344 & 6,895 & 3,397 \\
    PTB \cite{ptb_dataset} & 2 & 549 & 349 & 173 \\
    Ribeiro \cite{ribeiro2020automatic} & 7 & 827 & 554 & 273 \\
    \hline
  \end{tabular}
\end{table*}

Additionally, applying transfer learning directly to 1-D signals has shown encouraging results.
Strodthoff et al. \cite{ptbxl_analysis} reported significant improvements when pre-training \textit{xresnet1d101} \cite{xresnet} on the PTB-XL \cite{ptb_dataset} dataset and subsequently fine-tuning on smaller datasets.
Weimann et al. \cite{weimann2021transfer} achieved up to a 6.57\% improvement in the classification performance of Atrial Fibrillation using CNNs, pre-trained on the large Icentia11K dataset \cite{tan2019icentia11k}.
Jang et al. \cite{jang2021effectiveness} showed that pre-training a convolutional autoencoder on the AUMC ICU dataset \cite{lee2016constructing} of size 26,481 worked better than training from random initialization on the 10,646-sample dataset of the Shaoxing People's Hospital of China \cite{chapman-shaoxing}.
Other studies have also reported positive results of transfer learning \cite{chen2019transfer,ghaffari2019atrial,li2012detecting,melep,jin2022transfer,mohebbanaaz2022new}.

While previous 1-D approaches have demonstrated the effectiveness of transfer learning in ECG diagnosis, these studies often focused on specific datasets and model architectures. An implicit assumption is that transferring knowledge from a large upstream dataset consistently improves downstream performance on another dataset, compared to training from random initialization (scratch). 
However, this hypothesis has not been systematically verified.
In this study, we aim to validate the hypothesis by testing it across different ECG datasets and deep learning architectures.
Specifically, we conduct extensive experiments using three upstream datasets for pre-training models and five downstream datasets for fine-tuning pre-trained models. 
We employ six deep learning models, encompassing the two predominant architectures for ECG diagnosis: Convolutional Neural Networks \cite{cai2024one,fan2018multiscaled,li2018patient,wang2020deep,petmezas2021automated,baloglu2019classification,guo2023ecg,limam2017atrial,loh2023deep} and Recurrent Neural Networks (RNNs) \cite{singh2018classification,prabhakararao2020attentive,kumar2022deepaware,saadatnejad2019lstm,limam2017atrial,petmezas2021automated,gutierrez2024non,faust2018automated}. The comparison between fine-tuning performance and training from scratch provides insights into the effectiveness of transfer learning in ECG applications. 
Our key contributions and findings are as follows:

\begin{itemize}
  \item We conduct the first extensive study on the effectiveness of transfer learning in the ECG domain, including six popular DNN architectures and five ECG datasets.
  \item Contrary to expectations, fine-tuning does not consistently outperform training from scratch. Its advantages diminish as the size of the downstream dataset increases.
  \item Fine-tuning can accelerate convergence, whereas training from scratch generally requires a longer time to sufficiently converge.
  \item For ECG data, fine-tuning demonstrates greater effectiveness with CNNs than with RNNs.
\end{itemize}

\section{Materials \& Methods}

\subsection{Datasets}

We used five publicly available ECG datasets in this work.
The first was PTB-XL \cite{ptbxl_dataset}, containing 21,837 ECG records from 18,885 patients, covering 44 diagnostic statements. 
Signals were sampled at either 500 Hz or 1000 Hz, with a duration of ten seconds each.
The 44 labels were categorized into five superclasses, namely: NORM (normal ECG), MI (Myocardial Infarction), STTC (ST/T-Changes), HYP (Hypertrophy), and CD (Conduction Disturbance) \cite{ptbxl_analysis}.
We focused on these five superclasses when conducting experiments with this dataset.

The second dataset was from the China Physiological Signal Challenge 2018 (CPSC2018) \cite{cpsc2018}, including 6,877 ECG records, sampled at 500 Hz and lasted for 6-60 seconds each. There are nine diagnostic labels: NORM, AF (Atrial Fibrillation), I-AVB (First-degree atrioventricular block), LBBB (Left Bundle Branch Block), RBBB (Right Bundle Branch Block), PAC (Premature Atrial Contraction), PVC (Premature ventricular contraction), STD (ST-segment Depression), and STE (ST-segment Elevated).

The third was the Georgia dataset \cite{georgiadataset}, consisting of 10,344 ECG signals with 10 seconds in length and a sampling rate of 500 Hz.
The dataset has a diverse range of 67 unique diagnoses.
However, our research concentrated on a subset of 10 specific labels having the highest number of samples: 
NORM, AF, I-AVB, PAC, SB (Sinus Bradycardia), LAD (left axis deviation), STach (Sinus Tachycardia), TAb (T-wave Abnormal), TInv (T-wave Inversion), and LQT (Prolonged QT interval).

The fourth was the PTB Diagnostic ECG Database \cite{ptb_dataset}, containing 549 ECG records sampled at 1000 Hz. We focused on two diagnostic classes: Myocardial Infarction (MI) and Healthy controls (NORM), covering 200 over 268 subjects involved in this dataset(it is worth noting that while there are ECG records from 290 subjects, clinical summaries are available for only 268 of them).

The last source was the Ribeiro dataset \cite{ribeiro2020automatic}. This contains 827 ECG records with seven annotations: NORM, I-AVB, RBBB, LBBB, SB, AF, and STach.

We reduced the sampling frequency of all ECG records to 100 Hz. This helps reduce computational load while retaining essential information.
In addition, all records need to have the same duration. Since most ECG signals in the five datasets lasted for ten seconds, we used this as the desired duration. 
For records exceeding this timeframe, we applied cropping.
For shorter records, since they only account for a tiny fraction, specifically six out of 6,877 records in the CPSC2018 dataset and 52 out of 10,334 records in the Georgia dataset, we simply omitted them. Each dataset was then split into training and test subsets with a test size ratio of 0.33.

\begin{figure*}[t]

  \begin{minipage}[t]{0.27\linewidth}
    \centering
    \centerline{\includegraphics[width=6cm]{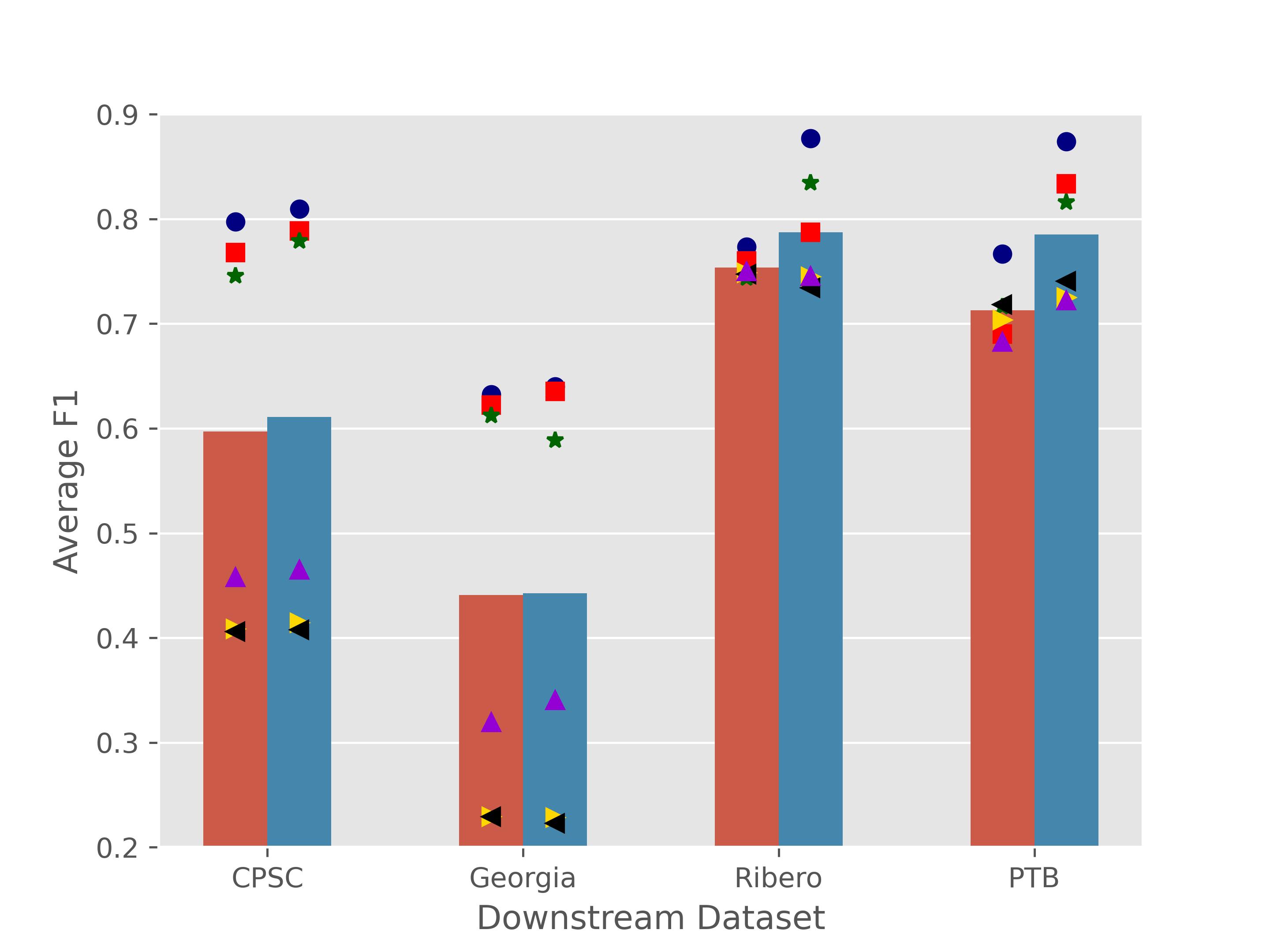}}
   \label{fig:bar_up_ptbxl}
    \centerline{\footnotesize (a) Pre-trained on PTB-XL}\medskip
  \end{minipage}
  \hfill
  \begin{minipage}[t]{.27\linewidth}
    \centering
    \centerline{\includegraphics[width=6.0cm]{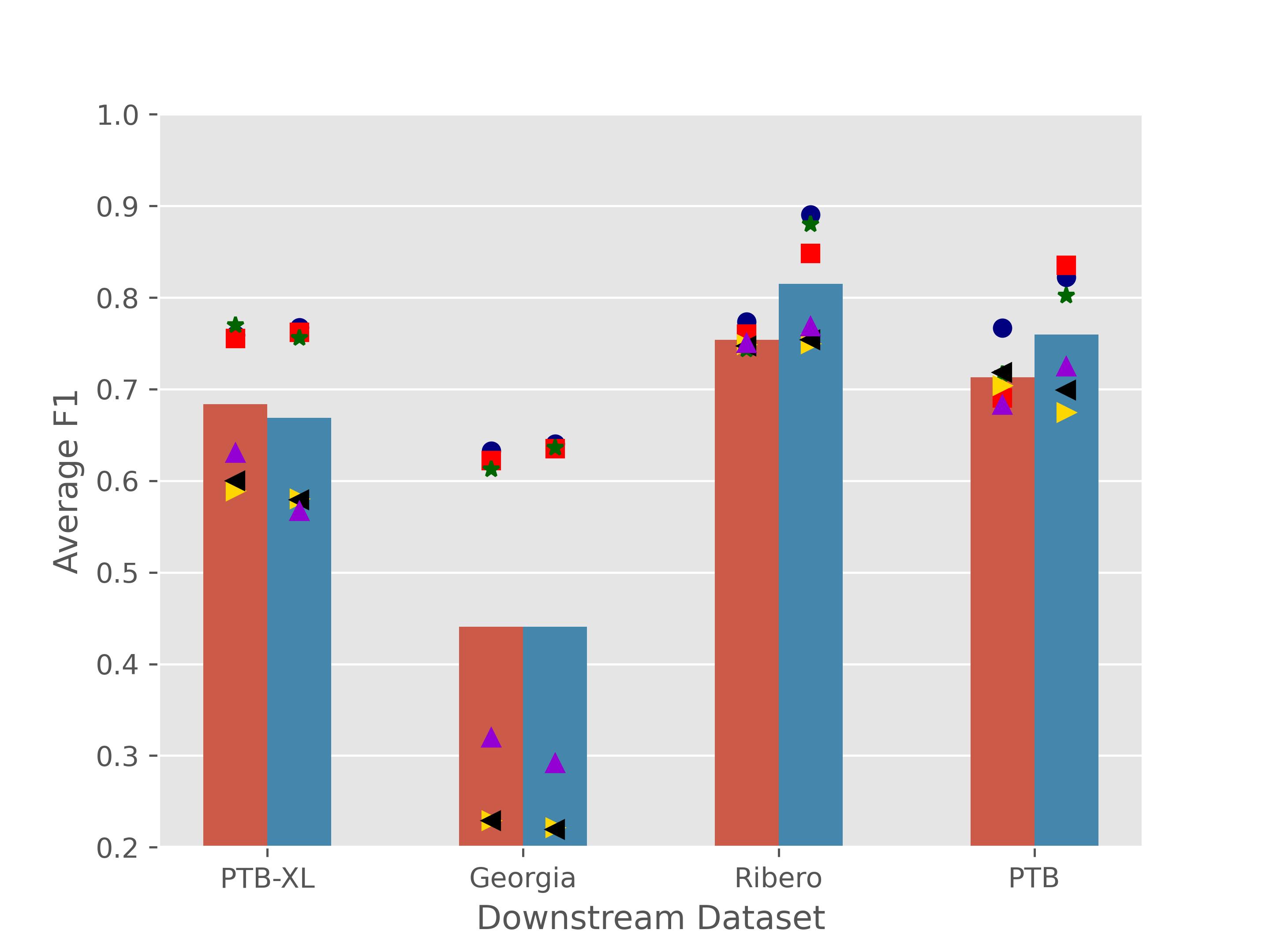}}
    \label{fig:bar_up_cpsc}
    \centerline{\footnotesize (b) Pre-trained on CPSC2018}\medskip
  \end{minipage}
  \hfill
  \begin{minipage}[t]{0.3\linewidth}
    \centering
    \centerline{\includegraphics[width=7.1cm]{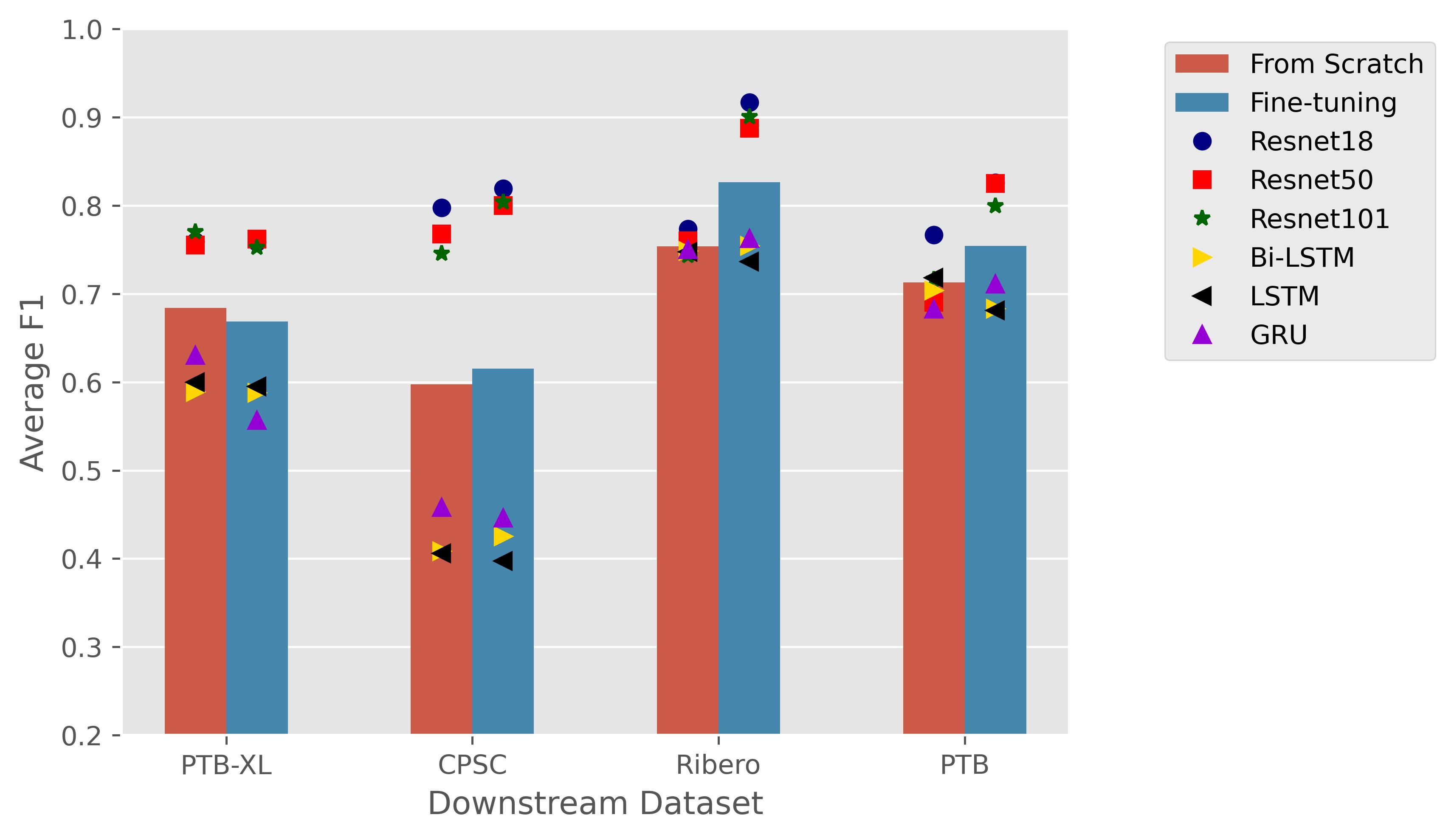}}
    \label{fig:bar_up_georgia}
    \centerline{ \footnotesize \hspace{-2cm} (c) Pre-trained on  Georgia}\medskip
  \end{minipage}
  \caption{Performance comparison of fine-tuning and training from scratch, with three upstream datasets, six models, and four downstream datasets. In each chart, six symbols depict the average $f_1$-scores for the respective models, and the bar shows the mean average score across these six models.}
  \label{fig:bar_chart}

  \begin{minipage}[t]{0.3\linewidth}
    \centering
    \centerline{\includegraphics[width=8cm]{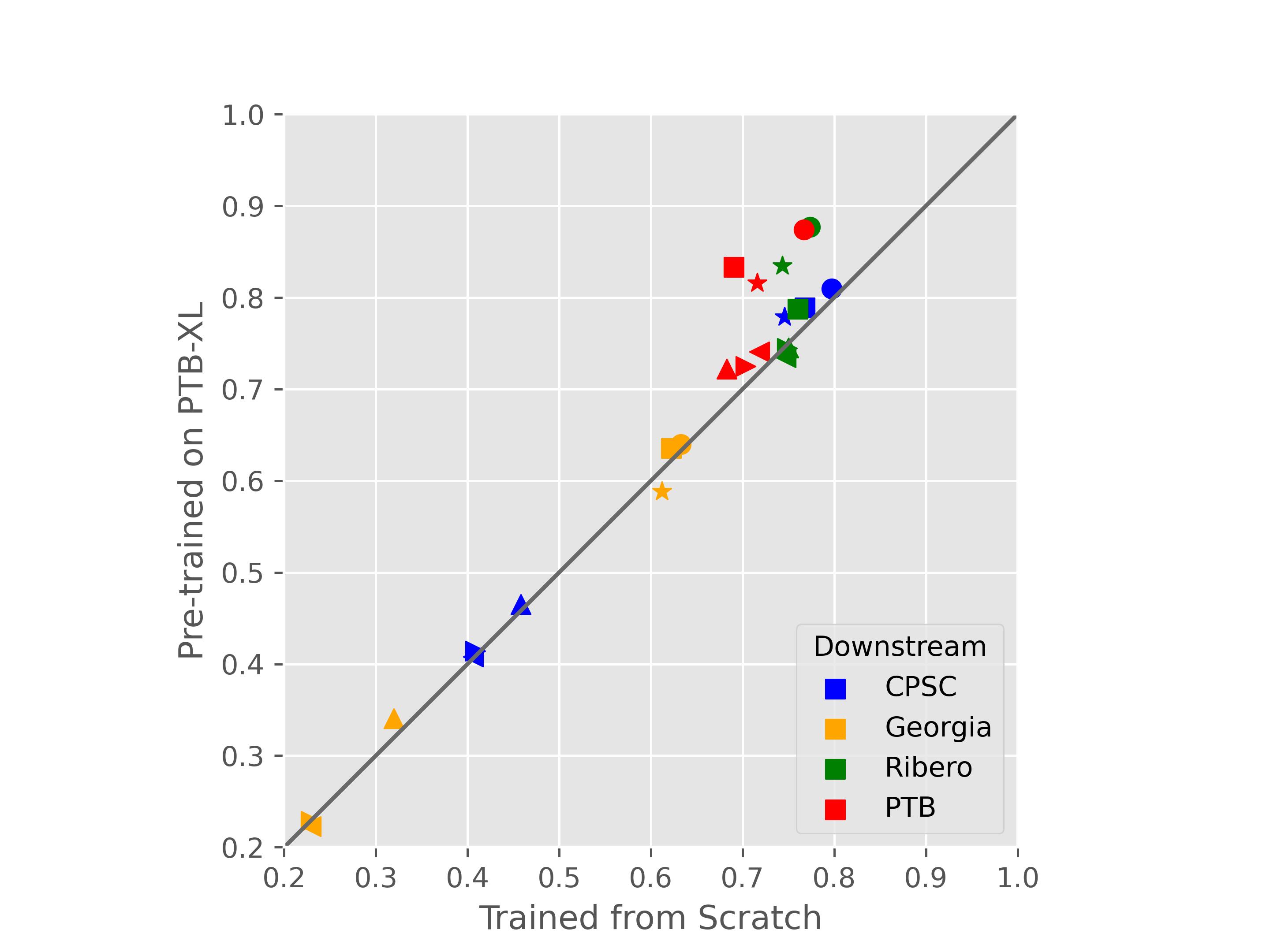}}
    \centerline{\footnotesize (a) Pre-trained on PTB-XL}\medskip
  \end{minipage}
  \hfill
  \begin{minipage}[t]{.3\linewidth}
    \centering
    \centerline{\includegraphics[width=8.0cm]{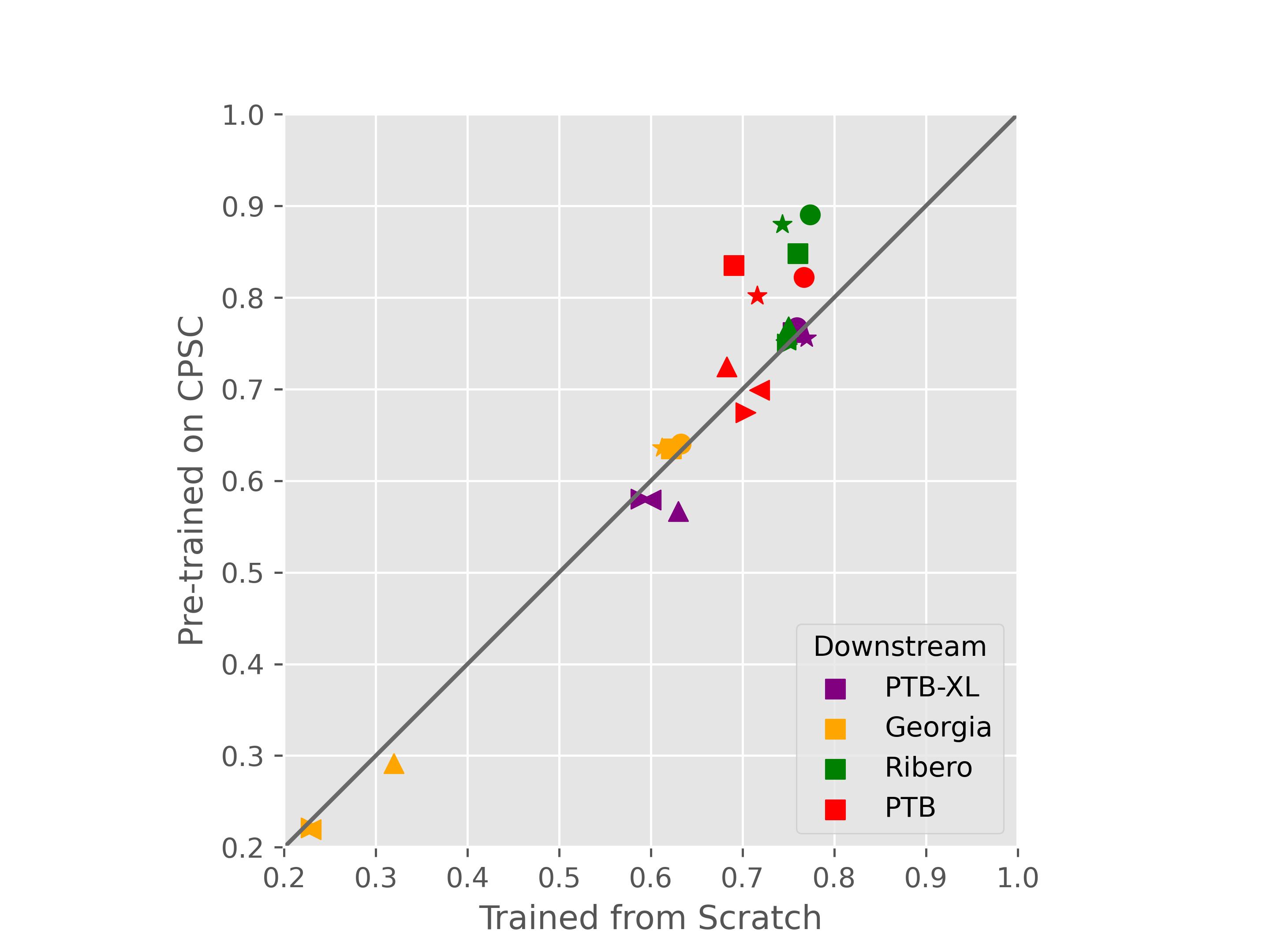}}
    \centerline{\footnotesize (b) Pre-trained on CPSC2018}\medskip
  \end{minipage}
  \hfill
  \begin{minipage}[t]{0.3\linewidth}
    \centering
    \centerline{\includegraphics[width=8cm]{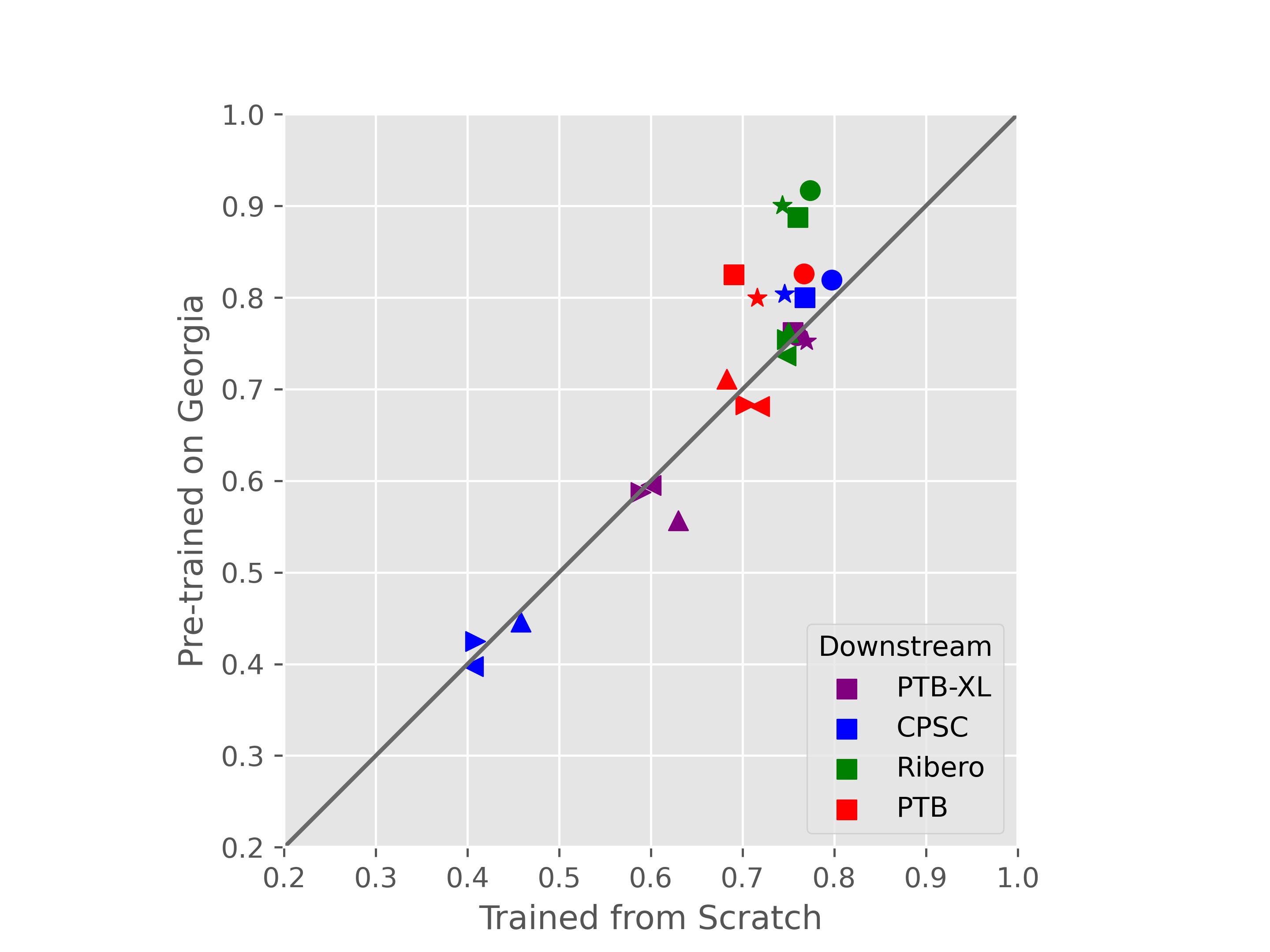}}
    \centerline{ \footnotesize (c) Pre-trained on  Georgia}\medskip
  \end{minipage}
  \caption{Another view of average-$f_1$ comparison between fine-tuning (vertical axis) and training from scratch (horizontal axis). Each point corresponds to a specific model and downstream dataset combination. Model legend is the same as in Fig. \ref{fig:bar_chart}. Best viewed in color. That the majority of points lying above the identity line suggests that fine-tuning generally outperformed training from scratch. However, this is not always true.}
  \label{fig:scatter_plot}
\end{figure*}

\subsection{Experiment Settings}

\subsubsection{Evaluation Metric}
We evaluated model performance on a dataset using their average \f1 on the test subset across all labels, weighted by the number of samples belonging to each label in the test subset. Importantly, the test subset was only used for evaluation during each training epoch and was never employed to update the model's parameters, ensuring prevention of data leakage \cite{kaufman2012leakage}.

\subsubsection{Pre-training}
\label{pretraining}
In this work, we examined six DNN architectures.
Three of these were convolutional: ResNet1d18, ResNet1d50, and ResNet1d101, which were adapted from the original 2-D versions \cite{resnet}.
The other three were recurrent DNNs: Long Short Term Memory (LSTM) \cite{lstm}, Bidirectional LSTM \cite{lstm}, and Gated Recurrent Unit (GRU) \cite{gru}.
Three datasets were used for pre-training: PTB-XL, CPSC2018, and Georgia, due to their substantial sample sizes.
Each of the six models was pre-trained on the training subset of each dataset for 100 epochs using Adam optimizer \cite{adam} with learning rate $0.01$.
We evaluated each model on the test subset during training, and only the checkpoint that achieved the best evaluation metric over 100 epochs was saved as the pre-trained model.
We opted not to save the last checkpoint at the 100$^{th}$ epoch due to observed overfitting as training progressed, especially for the three recurrent models. 

Training a model from scratch involved the same process described above and was applied to all five datasets, not limited to the three largest ones.

\subsubsection{Fine-tuning}

When fine-tuning a pre-trained model on a downstream dataset, as the number of output neurons may be different, we replaced the top fully-connected layer in the pre-trained model with a new layer with the number of neurons equal to the number of labels in the downstream dataset. 
For example, when fine-tuning ResNet1d18, which was pre-trained on PTB-XL (five labels), on Ribeiro as the downstream dataset (seven labels), we replaced the top layer with five outputs with a new one with seven outputs, and kept the layer's input unchanged.
Then the whole model underwent the same training procedure as pre-training, described in Section \ref{pretraining}.

\section{Results}

\begin{figure}[h]
  \centering
  \includegraphics[width=\linewidth]{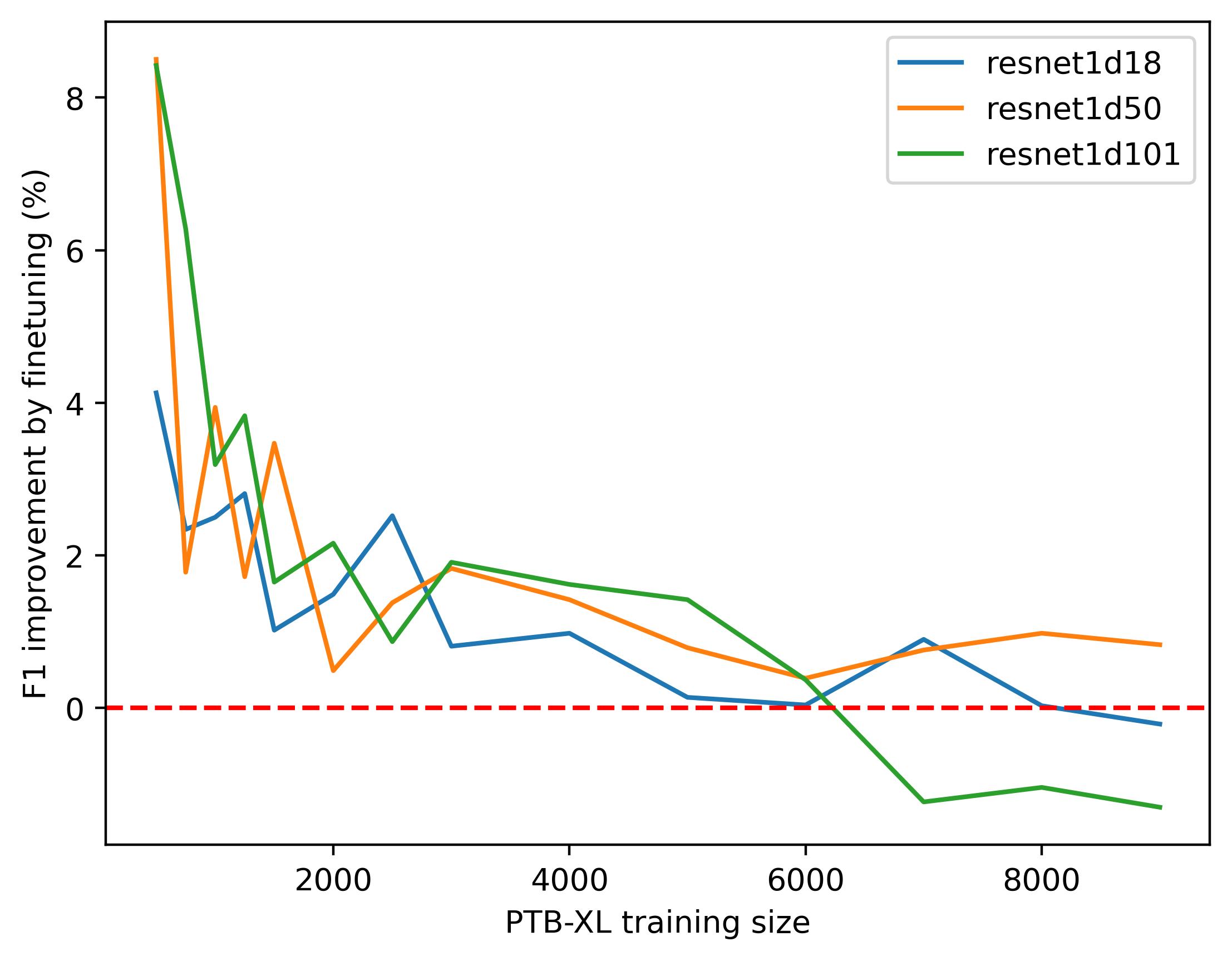}
\caption{Fine-tuning improvement of the three ResNets with varying downstream dataset size.}
\label{fig:ptbxl_varying_size}
\end{figure}

\subsection{Fine-tuning does not necessarily improve performance}
\label{sec:compare_finetune}

Figure \ref{fig:bar_chart} illustrates the performance comparison between fine-tuning and training from scratch. 
Each chart corresponds to one of the three upstream datasets, with the results of all six models on each downstream dataset scattered for both cases.
The bars denote the average performance across the six models, providing the overall comparison.

Clearly, transfer learning does not consistently outperform training from scratch.
On one hand, it significantly improved the model's performance on Ribeiro and PTB, the two small downstream datasets.
On the other hand, when using Georgia as the downstream dataset, there is little average difference in performance, despite variations among individual models. This is depicted in Figure \ref{fig:bar_chart}(a) for PTB-XL and Figure \ref{fig:bar_chart}(b) for CPSC2018 as upstream datasets.
Notably, when fine-tuning on PTB-XL, the overall performance is slightly poorer than training from random initialization, regardless of whether pre-training occurred on CPSC2018 or Georgia, as seen in Figure \ref{fig:bar_chart}(b) and \ref{fig:bar_chart}(c).

Figure \ref{fig:scatter_plot} shows an alternative perspective on the comparison, using the same model legend as in Figure \ref{fig:bar_chart}.
Each point on the plot represents a model and downstream dataset combination. 
In the scenario of pre-training on PTB-XL (Figure \ref{fig:scatter_plot}(a)), nearly all points remained above the identity line, indicating the superior performance of fine-tuning.
However, after pre-training on CPSC2018 and Georgia datasets, fine-tuning RNNs mostly led to poorer results, as more triangular symbols (representing RNNs) fell below the line, especially for PTB-XL as the downstream dataset (shown as purple points).
This observation aligns with the results shown in Figure \ref{fig:bar_chart}.

\begin{figure*}[t]
  \centering
  \includegraphics[width=\linewidth]{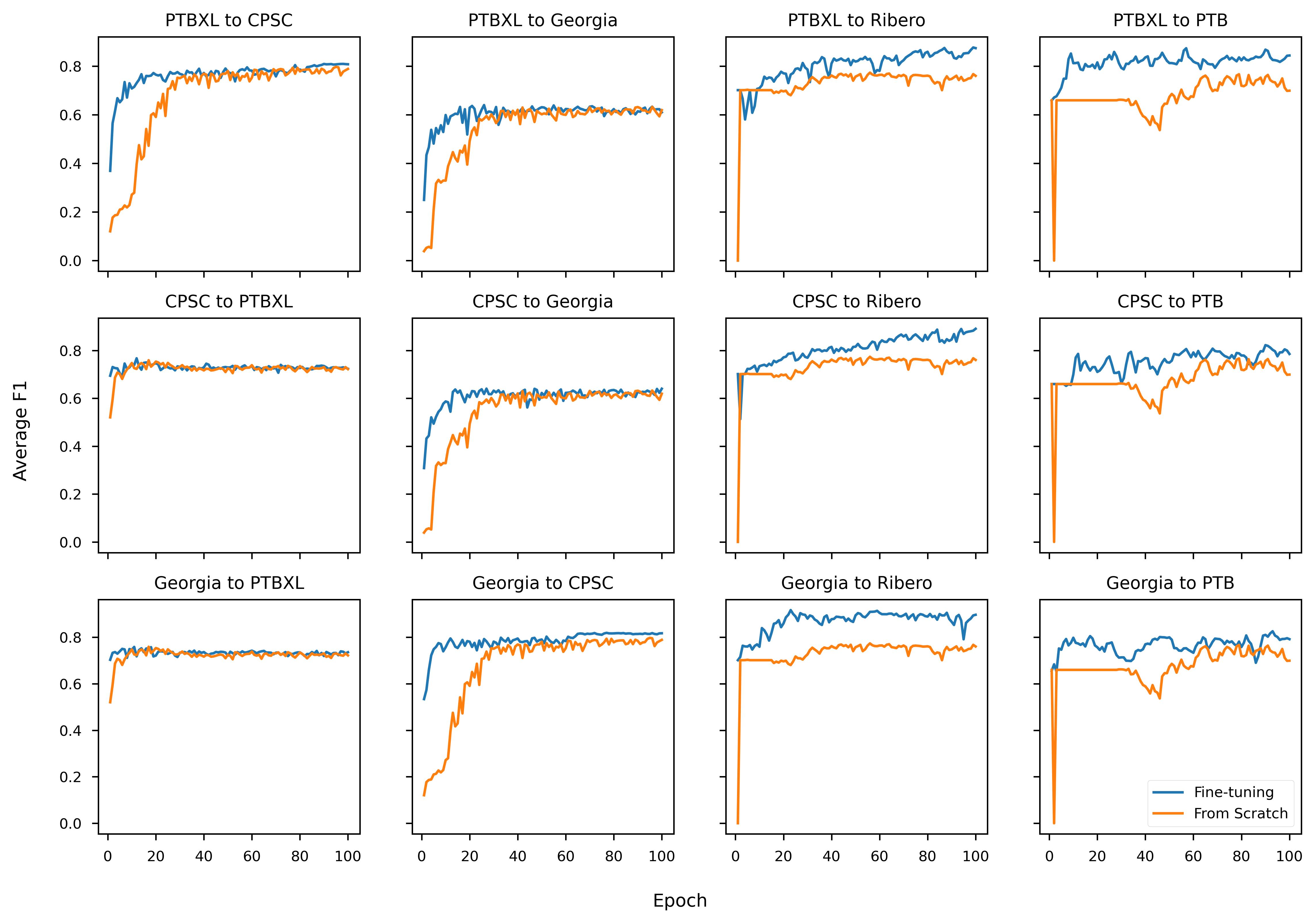}
\caption{Performances of ResNet1d18 during fine-tuning and training from scratch. Three rows represent three upstream datasets: PTB-XL, CPSC2018, and Georgia, respectively.}
\label{fig:resnet18_epoch}
\end{figure*}

\subsection{Fine-tuning improvement fades with downstream dataset size}

The results presented in Section \ref{sec:compare_finetune} suggest that the comparison between fine-tuning and training from scratch is influenced by the size of the downstream dataset. Fine-tuning exhibited the most significant improvement over training from scratch when the dataset size was small (as observed in the cases of Ribeiro and PTB), with diminishing improvement as larger datasets (PTB-XL, CPSC2018, and Georgia) were used.
To gain better insights, we conducted experiments with three pre-trained ResNets on the Georgia dataset.
For the downstream task, we varied the size of the PTB-XL training set from 500 to 9000 samples, measuring the average \f1 improvement achieved by fine-tuning over training from scratch. To ensure a fair comparison, evaluation was conducted on the same PTB-XL test subset used in Section \ref{sec:compare_finetune}, regardless of the number of training samples.

Figure \ref{fig:ptbxl_varying_size} shows that performance gain through fine-tuning declined as the training size increased.
The most significant improvement occurred with a downstream dataset of 500 training samples, and training from scratch gradually reached comparable performance when the size reached 6000 samples.
Though fluctuations were present in the region of fewer than 2000 samples, likely because of the inherent randomness in deep learning \cite{raste2022quantifying}, overall the declining trend remains evident.
The results highlight the importance of transfer learning in the small dataset regime, though it may be less necessary when dealing with sufficiently large datasets.

\subsection{Fine-tuning can accelerate convergence}
\label{sec:accelerate_convergence}
While transfer learning might not consistently outperform training from scratch in terms of accuracy, the next question is whether it contributed to speeding up the training process.
We examined the evaluation metric across 100 epochs in all cases to answer this question.
Figure \ref{fig:resnet18_epoch} shows ResNet1d18's average \f1 at each training epoch on the corresponding downstream test subset.
Notably, in scenarios such as transferring from PTB-XL to CPSC2018, from PTB-XL to Georgia, from CPSC2018 to Georgia, and from Georgia to CPSC2018, although the performance of training from scratch eventually caught up with that of fine-tuning, it took approximately 30-35 epochs to do so.
Meanwhile, transferring from CPSC2018 to PTB-XL or from Georgia to PTB-XL offered minor accelerating benefits, and for small downstream datasets (Ribeiro, PTB), transfer learning was clearly superior.  
For results of other models, please refer to Appendix A.

\subsection{Fine-tuning tends to work better with CNNs than with RNNs}
\label{sec:compare_arch}

\begin{figure}[h]
  \centering
  \includegraphics[width=0.8\linewidth]{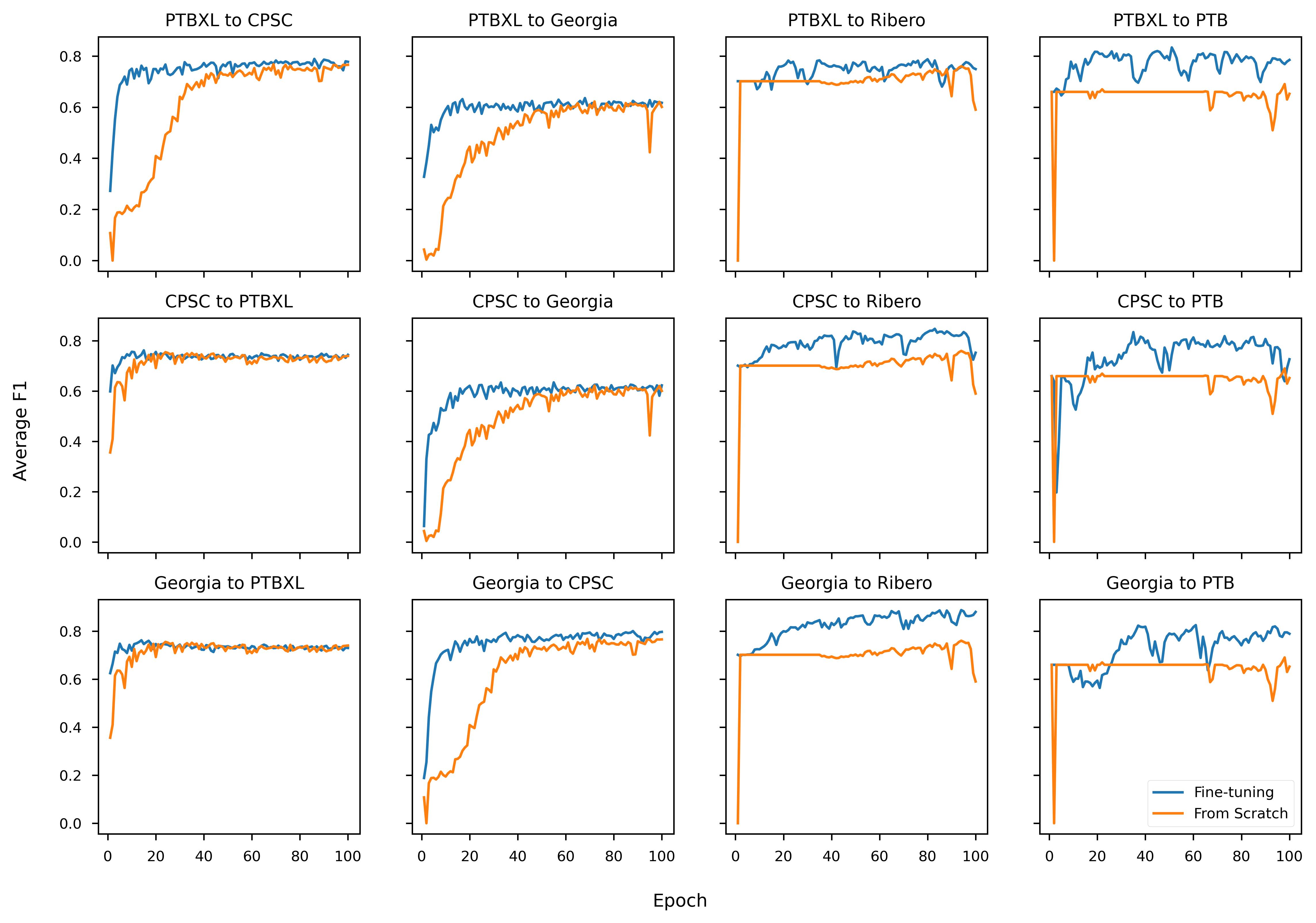}
\caption{Performances of ResNet1d50 during fine-tuning and training from scratch.}
\label{fig:resnet50_epoch}

\end{figure}

Concerning architectural selection, not only achieving higher overall \f1 than LTSM, Bi-LSTM, and GRU, three ResNet models (the circle, the square, and the star in Figure \ref{fig:bar_chart}) showed better compatibility with transfer learning. 
Fine-tuning those CNNs consistently resulted in improved performance compared to training from scratch in almost all scenarios, no matter which upstream and downstream datasets were used, as shown in Figure \ref{fig:bar_chart} and Figure \ref{fig:scatter_plot}.
In contrast, transfer learning had a minor impact on the three RNNs, as in numerous cases, their performance even lagged behind that of random initialization (see the up, left, and right triangles in the two figures). 
Moreover, when examining the convergence patterns (refer to Figure \ref{fig:resnet18_epoch} and Figures \ref{fig:resnet50_epoch},\ref{fig:resnet101_epoch},\ref{fig:lstm_epoch},\ref{fig:bilstm_epoch},\ref{fig:gru_epoch} in Appendix A), it is clear that fine-tuning CNNs played a crucial role in expediting and stabilizing the convergence process, whereas RNNs (especially GRU) exhibited a notably more erratic result.

This phenomenon can be explained by the inherent characteristics of the two architectures.
Convolutional layers within CNNs are adept at capturing spatial features such as shapes, patterns, peaks, and troughs—features that are low-level and do not necessitate relearning during fine-tuning on downstream datasets
On the other hand, LTSM and GRU specialize in capturing temporal dependencies, processing signals sequentially to maintain a "memory" that is high-level and complex. 
Consequently, the learned memory from one dataset may not be applicable or effective for others, rendering the transfer of such memory ineffective.
Furthermore, inherent challenges in training RNNs, such as vanishing gradients \cite{vanishing_grad} and exploding gradients \cite{exploding_grad}, may exacerbate the difficulty of fine-tuning these networks.


\section{Conclusion}

In this work, we empirically investigate the effectiveness of transfer learning in multi-label ECG diagnosis through extensive experiments involving diverse datasets and deep learning models.
We show that when the downstream dataset is sufficiently large, pre-training may not exhibit superior performance compared to training from random initialization. This observation challenges the prevailing assumption that transfer learning invariably enhances performance across different tasks.
Nevertheless, in many real-world scenarios, the availability of small downstream datasets is a common constraint due to the substantial costs associated with data collection and annotation. In such cases, we assert that transfer learning remains a crucial and valuable approach.
Even when a decently large dataset is available, transfer learning will still be useful, as it can accelerate convergence, saving resources \& time and expediting both research and production cycles.

Moreover, our results confirm that fine-tuning tends to yield more effective results with CNNs than with RNNs in ECG classification. 
Contrary to 2-D images, RNNs are also a potential method to process time-series ECG signals. 
However, as mentioned in Section \ref{sec:compare_arch}, inherent designs of RNNs make it more difficult to transfer knowledge learned from one dataset to another.
Even in the case of training from scratch, LSTM, Bi-LSTM, and GRU showed inferior performance than that of ResNets (Section \ref{sec:compare_finetune}). 
Thus we argue that in general, CNNs should be the preferred choice when deciding on architectures for ECG applications.

\section*{Acknowledgement}
This research is supported by VinUni Seed Grant 2020.

\bibliographystyle{IEEEbib}
\bibliography{refs}

\clearpage
\onecolumn

\section*{Appendix A}

Supplementary materials for Section \ref{sec:accelerate_convergence}.

\begin{figure}[h]
\end{figure}

\begin{figure}[h]
  \centering
  \includegraphics[width=0.8\linewidth]{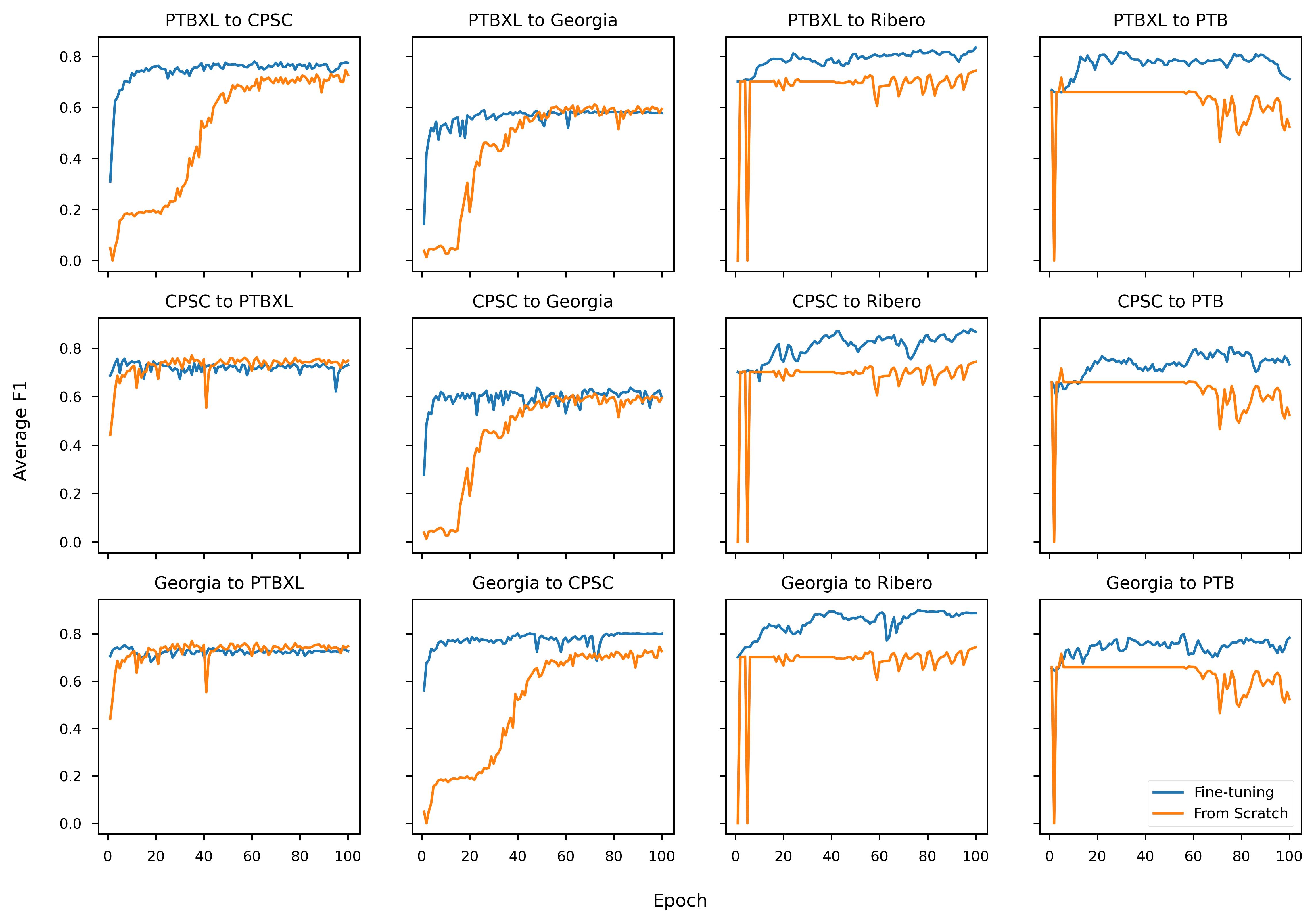}
  \caption{Performances of ResNet1d101 during fine-tuning and training from scratch.}
  \label{fig:resnet101_epoch}

  \bigskip

  \centering
  \includegraphics[width=0.8\linewidth]{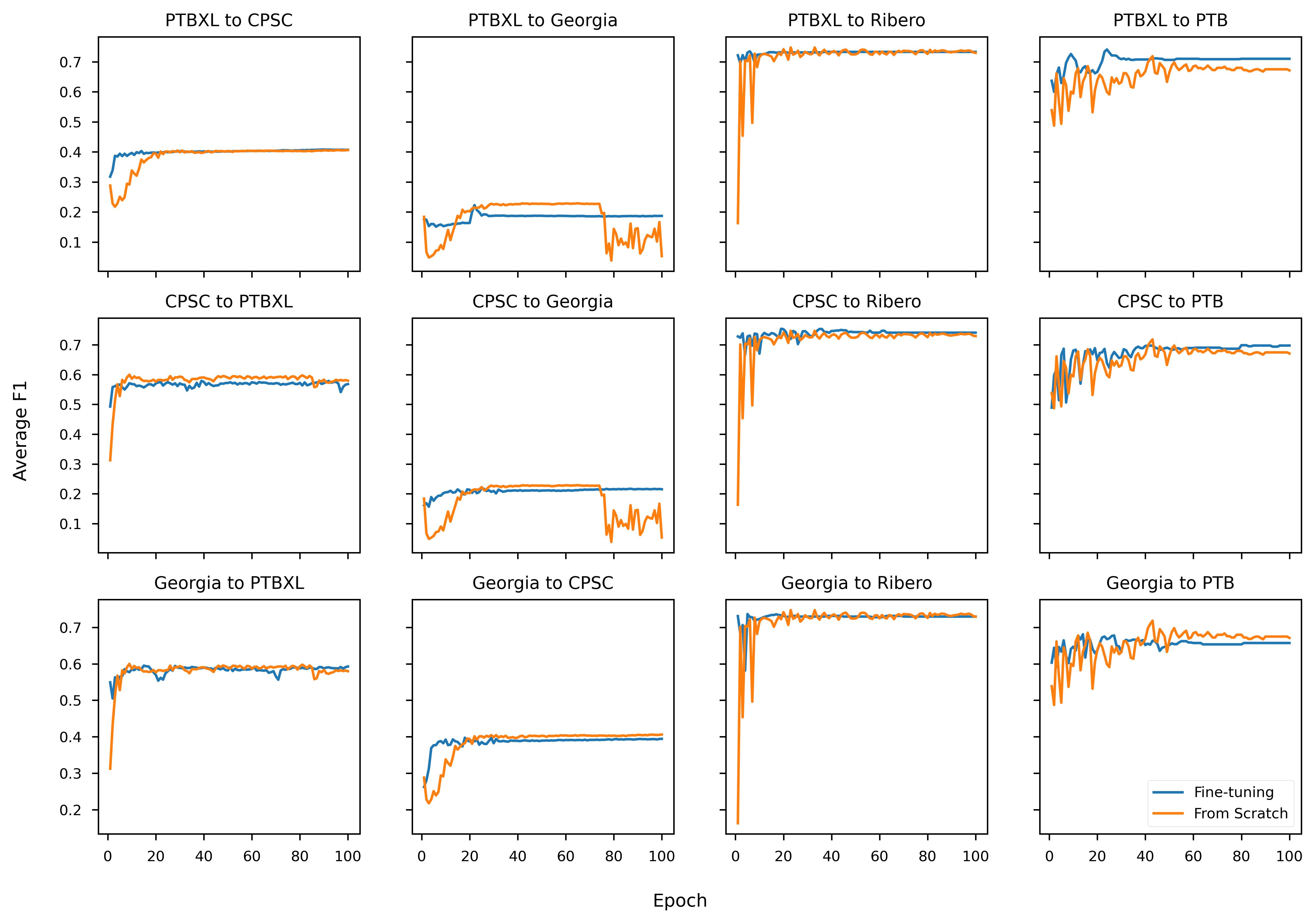}
\caption{Performances of LSTM during fine-tuning and training from scratch.}
\label{fig:lstm_epoch}
\end{figure}

\begin{figure}[h]
  \centering
  \includegraphics[width=0.8\linewidth]{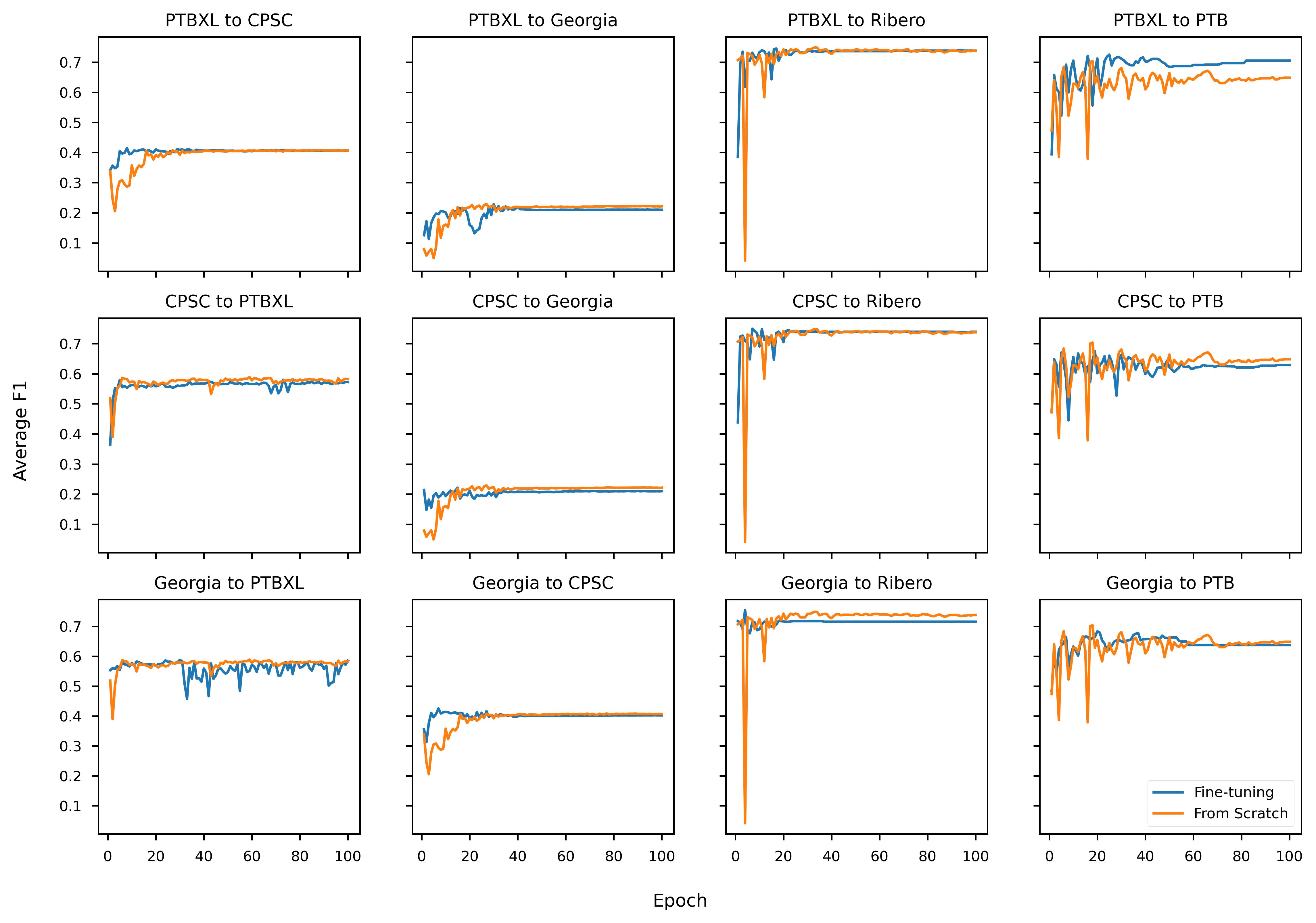}
\caption{Performances of Bi-LSTM during fine-tuning and training from scratch.}
\label{fig:bilstm_epoch}

\bigskip

\centering
\includegraphics[width=0.8\linewidth]{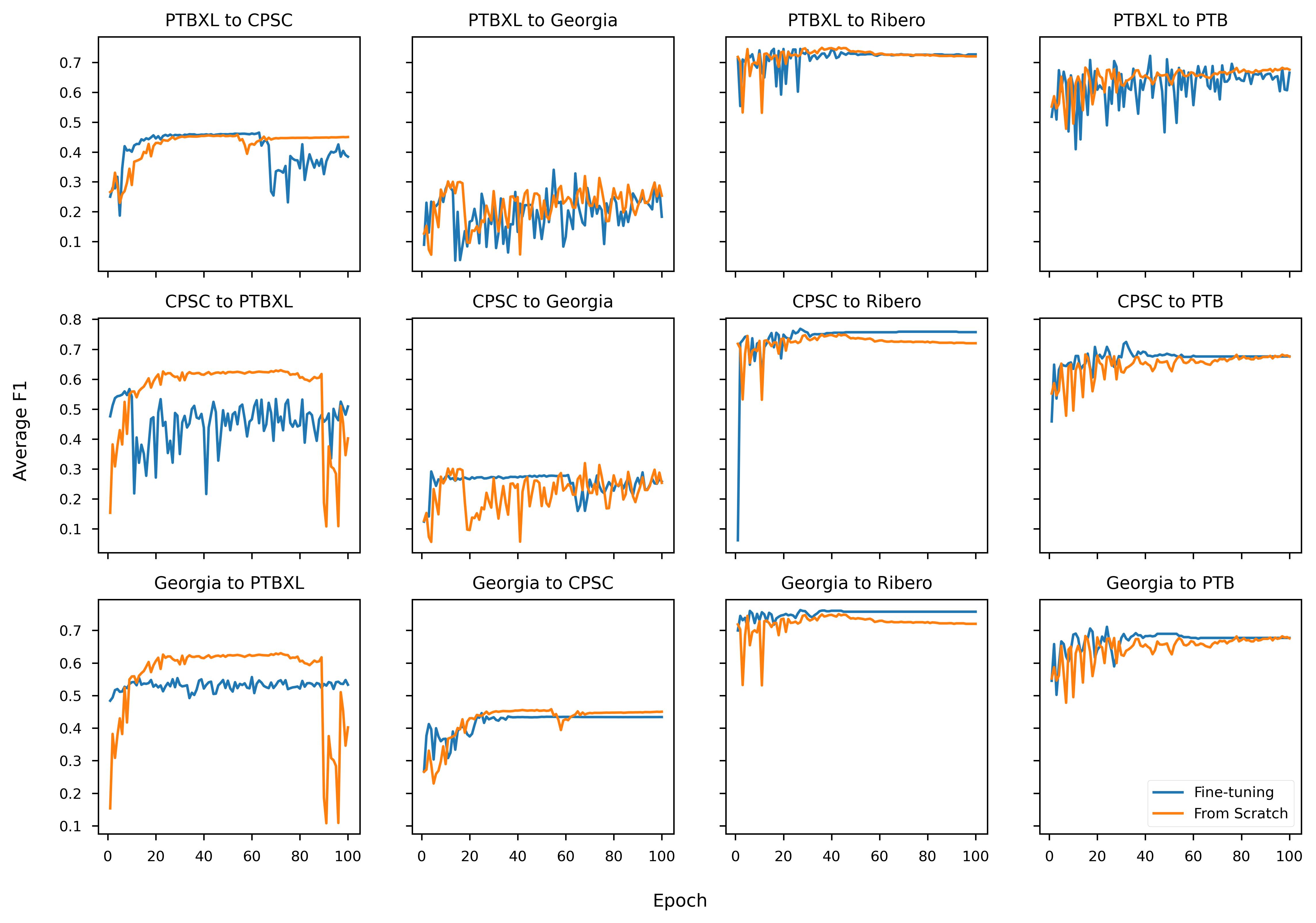}
\caption{Performances of GRU during fine-tuning and training from scratch.}
\label{fig:gru_epoch}
\end{figure}

\clearpage

\section*{Appendix B}
This section provides the full numerical results of our experiments in Section \ref{sec:compare_finetune} and \ref{sec:compare_arch}.

\begin{table*}[h]
  \centering
  \caption{Performance of training from scratch: maximum average \f1 during 100 epochs.} \vspace{2mm}
  \label{table:scratch_results}
  \begin{tabular}{|l|ccccc|}
    \hline
    & PTB-XL & CPSC & Georgia & Ribero & PTB \\
    \hline
    Resnet18 & 0.759 & 0.797 & 0.633 & 0.774 & 0.767 \\
    Resnet50 & 0.755 & 0.768 & 0.622 & 0.760 & 0.690 \\
    Resnet101 & 0.770 & 0.746 & 0.612 & 0.743 & 0.716 \\
    Bi-LSTM & 0.589 & 0.408 & 0.229 & 0.749 & 0.704 \\
    LSTM & 0.600 & 0.406 & 0.229 & 0.747 & 0.718 \\
    GRU & 0.630 & 0.458 & 0.319 & 0.750 & 0.683 \\
  \hline
\end{tabular}

\medskip

  \centering
  \caption{Fine-tuning performance of models pre-trained on PTB-XL: maximum average \f1 during 100 epochs.} \vspace{2mm}
  \label{table:up_ptbxl_results}
  \begin{tabular}{|l|cccc|}
    \hline
    & CPSC & Georgia & Ribero & PTB \\
    \hline
    Resnet18 & 0.809 & 0.640 & 0.877 & 0.874 \\
    Resnet50 & 0.789 & 0.636 & 0.787 & 0.833 \\
    Resnet101 & 0.779 & 0.589 & 0.834 & 0.816 \\
    Bi-LSTM & 0.414 & 0.228 & 0.745 & 0.725 \\
    LSTM & 0.408 & 0.223 & 0.734 & 0.741 \\
    GRU & 0.465 & 0.341 & 0.745 & 0.722 \\
\hline
\end{tabular}

\medskip

  \centering
  \caption{Fine-tuning performance of models pre-trained on CPSC: maximum average \f1 during 100 epochs.} \vspace{2mm}
  \label{table:up_cpsc_results}
  \begin{tabular}{|l|cccc|}
    \hline
    & PTB-XL & Georgia & Ribero & PTB \\
    \hline
    Resnet18 & 0.767 & 0.640 & 0.891 & 0.822 \\
    Resnet50 & 0.762 & 0.635 & 0.848 & 0.835 \\
    Resnet101 & 0.756 & 0.636 & 0.880 & 0.802 \\
    Bi-LSTM & 0.580 & 0.221 & 0.750 & 0.675 \\
    LSTM & 0.579 & 0.220 & 0.754 & 0.699 \\
    GRU & 0.567 & 0.292 & 0.768 & 0.724 \\
  \hline
\end{tabular}

\medskip

\centering
  \caption{Fine-tuning performance of models pre-trained on Georgia: maximum average \f1 during 100 epochs.} \vspace{2mm}
  \label{table:up_georgia_results}
  \begin{tabular}{|l|cccc|}
    \hline
    & PTB-XL & CPSC & Ribero & PTB \\
    \hline
    Resnet18 & 0.759 & 0.819 & 0.917 & 0.826 \\
    Resnet50 & 0.762 & 0.800 & 0.888 & 0.825 \\
    Resnet101 & 0.752 & 0.804 & 0.901 & 0.800 \\
    Bi-LSTM & 0.588 & 0.425 & 0.754 & 0.683 \\
    LSTM & 0.595 & 0.397 & 0.737 & 0.681 \\
    GRU & 0.557 & 0.446 & 0.762 & 0.711 \\
  \hline
\end{tabular}
\end{table*}

\end{document}